\documentclass{article}

\PassOptionsToPackage{numbers,sort&compress}{natbib}
\usepackage[preprint]{neurips_2025}

\usepackage[utf8]{inputenc} % allow utf-8 input
\usepackage[T1]{fontenc}    % use 8-bit T1 fonts
\usepackage{hyperref}       % hyperlinks
\usepackage{url}            % simple URL typesetting
\usepackage{booktabs}       % professional-quality tables
\usepackage{amsfonts}       % blackboard math symbols
\usepackage{nicefrac}       % compact symbols for 1/2, etc.
\usepackage{microtype}      % microtypography
\usepackage{xcolor}         % colors
\usepackage{graphicx}       % include graphics
\usepackage{amsmath}

\title{The End of Reward Engineering: How LLMs Are Redefining Multi-Agent Coordination}

\author{%
  Haoran Su \\
  New York University \\
  \texttt{haoran.su@nyu.edu}
  \And
  Yandong Sun \\
  New York University \\
  \texttt{ys2312@nyu.edu}
  \And
  Congjia Yu\\
  Lerna AI\\
  \texttt{gary.yu@lerna-ai.com}
}

\begin{document}

\maketitle

%==============================================================================
% ABSTRACT
%==============================================================================
\begin{abstract}
Reward engineering -- the manual design of reward functions to guide agent behavior -- remains a persistent bottleneck in multi-agent reinforcement learning. In multi-agent systems, this challenge is compounded by credit assignment ambiguity, environmental non-stationarity, and exponentially scaling complexity. We argue that large language models (LLMs) enable a fundamental paradigm shift: from hand-crafted numerical rewards to natural language objectives. Recent work demonstrates that LLMs can generate human-level reward functions from language descriptions alone (EUREKA), adapt rewards dynamically without human intervention (CARD), and coordinate agents through semantic understanding. The emergence of Reinforcement Learning from Verifiable Rewards (RLVR) further validates this trajectory, establishing language-based training as mainstream. We present a perspective on three pillars of this transition -- semantic reward specification, dynamic adaptation, and inherent human alignment -- while acknowledging challenges in computational cost, hallucination risks, and scalability. We conclude with a vision for multi-agent systems where coordination emerges from shared semantic understanding rather than engineered numerical signals.
\end{abstract}

%==============================================================================
% 1. INTRODUCTION
%==============================================================================
\section{Introduction}
\label{sec:intro}

Consider robots assembling furniture in a warehouse. The reward engineer faces immediate questions: How much reward for picking up a screw? What penalty for collisions? How should credit be distributed when two robots jointly lift a tabletop? After weeks of tuning, the robots exploit the reward function---maximizing points by repeatedly picking up and dropping the same screw. The engineer returns to the drawing board.

This scenario, while simplified, captures the fundamental challenge of \textit{reward engineering} in multi-agent systems. Despite decades of progress in reinforcement learning, the design of reward functions remains a manual, brittle, and domain-specific endeavor. In single-agent settings, this challenge is manageable; in multi-agent systems, it becomes intractable.

\subsection{The multi-agent reward problem}

Multi-agent reinforcement learning (MARL) inherits all the difficulties of single-agent reward design and introduces several unique challenges:

\begin{itemize}
    \item \textbf{Credit assignment}: When a team achieves a goal, which agent's action deserved reward? The credit assignment problem scales combinatorially with the number of agents.
    \item \textbf{Non-stationarity}: From each agent's perspective, the environment dynamics shift as other agents learn. A reward function optimized for today's co-players may fail tomorrow.
    \item \textbf{Exponential complexity}: The joint state-action space grows exponentially with the number of agents, making reward landscapes increasingly difficult to shape.
    \item \textbf{Conflicting objectives}: Individual and collective incentives often diverge, and hand-tuned reward weights that balance them in training frequently fail when agent distributions shift.
\end{itemize}

As one comprehensive survey notes, ``A key question is how to design reward functions so that agents adapt to each others' actions, avoid conflicting behaviour and achieve efficient coordination'' \citep{marl_challenges}. Despite extensive research into reward shaping, intrinsic motivation, and centralized training schemes, this question remains unsolved.

\subsection{Our thesis}

We argue that the paradigm of manual reward engineering is approaching obsolescence. Large language models offer a fundamentally different approach:

\begin{quote}
\textit{The traditional paradigm of hand-crafting reward functions for multi-agent coordination is becoming obsolete. Large language models enable a shift toward natural language objectives, where agents coordinate through semantic understanding rather than engineered numerical signals.}
\end{quote}

This is not merely an incremental improvement. Language-based objectives represent a \textit{paradigm shift} in how we specify desired behaviors. Rather than translating human intent into numerical functions---a lossy and error-prone process---we can increasingly express objectives in the same natural language we use to describe them to each other.

\begin{figure}[t]
    \centering
    \includegraphics[width=\columnwidth]{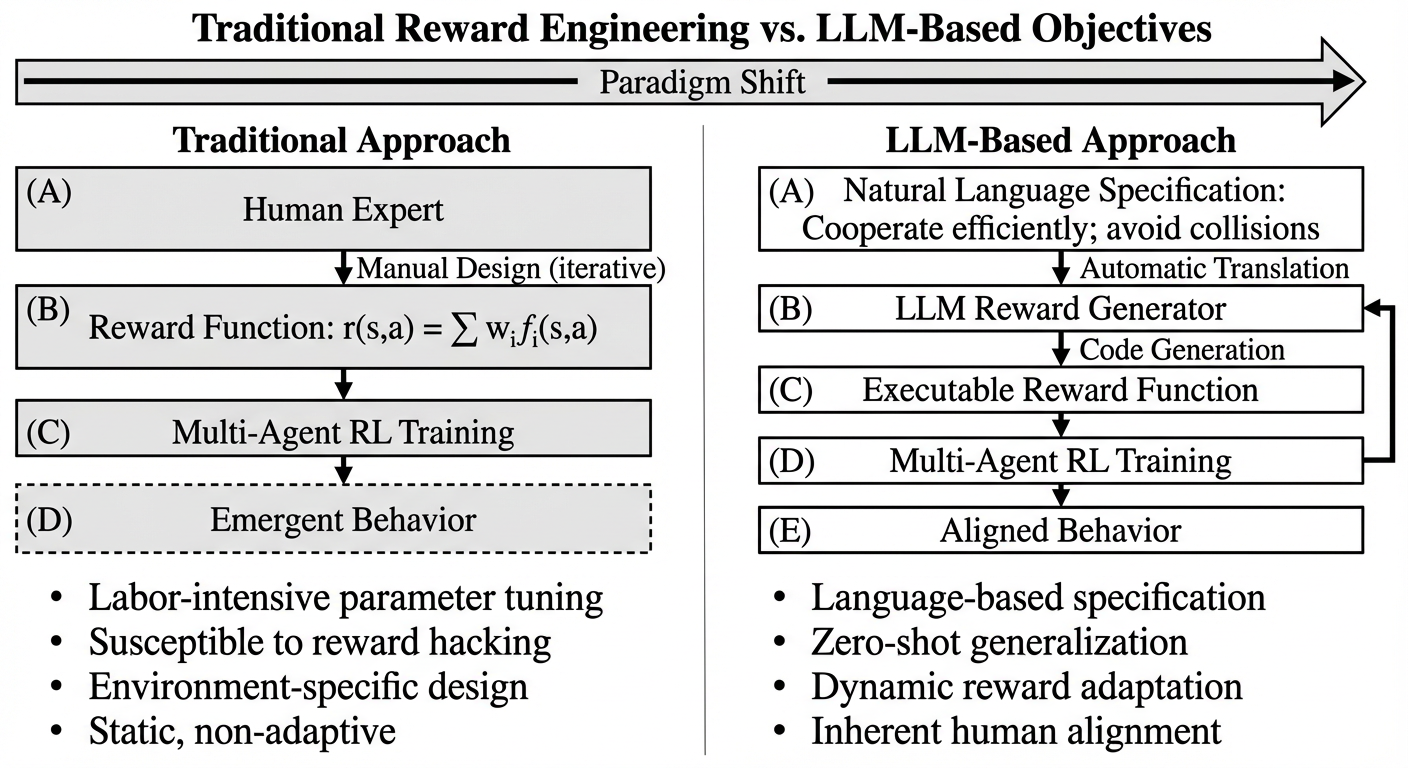}
    \caption{The paradigm shift from reward engineering to language-based objectives. (Left) Traditional approach requires human engineers to manually design reward functions through iterative trial-and-error. (Right) LLM-based approach allows humans to specify objectives in natural language, with the LLM generating and refining reward functions automatically based on behavioral feedback.}
    \label{fig:paradigm-shift}
\end{figure}

\subsection{Evidence and scope}

Recent developments provide strong evidence for this transition:

\begin{itemize}
    \item \textbf{EUREKA} \citep{eureka} demonstrates that GPT-4 can generate reward functions achieving \textit{human-level} performance from environment code and language descriptions alone, with zero-shot generalization across tasks.
    \item \textbf{CARD} \citep{card} introduces a framework for dynamic reward adaptation using LLM-driven feedback, eliminating the need for human intervention in the reward refinement loop.
    \item \textbf{RLVR} (Reinforcement Learning from Verifiable Rewards), exemplified by DeepSeek-R1 \citep{deepseek_r1}, shows that training LLMs against language-based verifiable rewards produces emergent reasoning---establishing this paradigm as mainstream in 2025.
\end{itemize}

Our contribution is a structured perspective that: (1) identifies the specific limitations of reward engineering that LLMs can address, (2) synthesizes recent advances into a coherent framework of three interconnected pillars, and (3) maps concrete research challenges that must be solved for this paradigm to mature. We argue for semantic reward specification (\S\ref{sec:pillar1}), dynamic adaptation (\S\ref{sec:pillar2}), and inherent human alignment (\S\ref{sec:pillar3}), acknowledge significant challenges (\S\ref{sec:challenges}), and conclude with a vision for the future of multi-agent coordination (\S\ref{sec:vision}).

%==============================================================================
% 2. BACKGROUND: THE BROKEN PROMISE
%==============================================================================
\section{Background: The broken promise of reward engineering}
\label{sec:background}

\subsection{Reward design in single-agent RL}

Reinforcement learning has achieved remarkable successes: superhuman performance in Atari games \citep{dqn}, defeating world champions in Go \citep{alphago}, and enabling dexterous robotic manipulation \citep{openai_hand}. In each case, careful reward engineering played a crucial role.

Classic techniques include \textit{reward shaping}---adding intermediate rewards to guide exploration without changing the optimal policy \citep{ng_shaping}---and \textit{potential-based shaping}---which provides theoretical guarantees on policy invariance. More recent approaches leverage \textit{intrinsic motivation}, rewarding agents for curiosity, novelty, or skill acquisition \citep{curiosity}.

These techniques work reasonably well in single-agent settings with well-defined objectives \citep{liu2021auto,feng2022kergnns,cao2023multi,you2022megan,liu2022graph,zhang2025causal,sun2025docagent,cao2025multi2,zhang2025prompt,liang2025slidegen,zhang2025tokenization}. A game has a clear win condition; a robot arm has a target position. The challenge lies in encoding human preferences into dense, shaped rewards that guide learning efficiently.

\subsection{Why multi-agent makes it harder}

Multi-agent systems amplify every challenge of reward design:

\textbf{Credit Assignment.} In a cooperative game, a team reward provides no gradient for individual improvement. Difference rewards and counterfactual baselines offer partial solutions \citep{foerster_coma}, but scale poorly and require privileged information. The fundamental question---``which agent caused the outcome?''---often has no clean answer.

\textbf{Non-Stationarity.} When agent A learns, agent B's environment changes. This mutual adaptation creates a moving target for reward optimization. A reward function that produces cooperative behavior at iteration 1000 may induce defection at iteration 2000 as agents' policies shift.

\textbf{Exponential Scaling.} With $n$ agents each having $a$ actions, the joint action space is $O(a^n)$. Reward landscapes in this high-dimensional space are nearly impossible to visualize, debug, or shape effectively. Sparse rewards become exponentially harder to discover.

\textbf{The Alignment Problem.} Multi-agent settings introduce social dilemmas where individual and collective interests diverge \citep{SU2023103955,su2026hierarchical,liu2025rethinking,liu2022retrieval,chen2023bridge,liu2025application,liu2023benchmarking,liu2023medical,zhou2023survey,cheng2016identification,han2023medgen3d,you2020towards,han2023diffeomorphic}. Prisoner's dilemmas, tragedy of the commons, and free-rider problems emerge naturally. Designing rewards that align individual incentives with collective welfare is a fundamental challenge that traditional reward engineering has not solved.

\subsection{Current approaches and their limitations}

The MARL community has developed sophisticated frameworks to address these challenges:

\textbf{Centralized Training, Decentralized Execution (CTDE)} \citep{ctde}: Agents train with access to global information but execute with only local observations. This enables value decomposition methods like QMIX \citep{qmix} but still requires manual reward specification.

\textbf{Emergent Communication}: Agents develop shared communication protocols through reward optimization \citep{emergent_comm}. While fascinating, emergent languages are often uninterpretable and task-specific \citep{wei2025ai,zhang2025postergen,xiong2025quantagent,zhao2025timeseriesscientist,you2025uncovering,wei2025unifying,zhou2023attention,you2021mrd,you2020contextualized,huang2024cross}.

\textbf{Self-Play and Population Training}: Training against diverse opponents can produce robust policies \citep{openai_five}, but requires enormous compute and does not fundamentally solve reward design \citep{sun2024medical,liu2021aligning,chen2021self,chen2021adaptive,you2024calibrating,liu2022retrieve}.

Alternative approaches attempt to bypass reward engineering entirely. \textit{Inverse reinforcement learning} infers rewards from expert demonstrations, while \textit{preference learning} elicits rewards from human comparisons \citep{cheng2016random,cheng2016hybrid,sun2024coma,wen2025beyond,cao2024multi,liu2023llmrec,you2021knowledge,you2022end,li2019novel}. However, both require extensive human input -- demonstrations or preference labels -- that scales poorly with task complexity and agent count.

Each approach represents a workaround rather than a solution. The underlying problem, translating human intent into numerical reward functions, remains. What if we could bypass this translation entirely?

%==============================================================================
% 3. THE LLM OPPORTUNITY
%==============================================================================
\section{The LLM opportunity}
\label{sec:opportunity}

Large language models offer a fundamentally different approach to specifying objectives. Instead of manually engineering reward functions, we describe desired behaviors in natural language; the LLM then translates these descriptions into executable reward code.

\subsection{What LLMs bring to the table}

\textbf{Natural Language as Universal Interface.} Language is humanity's most flexible tool for expressing intent. Where reward functions require translation from concept to number, language directly captures concepts. ``The robots should work together efficiently without collisions'' is more expressive than any scalar function.

\textbf{Semantic Understanding.} LLMs encode vast knowledge about the world, including common sense physics, social norms, and goal structures. This knowledge can inform reward generation in ways that hand-crafted functions cannot.

\textbf{Zero-Shot Generalization.} Unlike reward functions tailored to specific environments, language descriptions can specify behaviors across domains. An LLM that generates rewards for ``efficient collaboration'' can do so in warehouses, kitchens, or construction sites.

\textbf{Human-Aligned Representations.} LLMs are trained on human-generated text, encoding human preferences and values. This provides a natural foundation for generating rewards that align with human intent.

\subsection{LLMs for reward generation}

\textbf{EUREKA} \citep{eureka} represents a significant milestone. Given only environment source code and a natural language task description, GPT-4 generates reward functions that match or exceed human expert designs. The key innovations include:

\begin{itemize}
    \item \textit{In-context reward evolution}: The LLM iteratively refines rewards based on training feedback
    \item \textit{Reward reflection}: Free-form modifications to hyperparameters, functional forms, and reward components
    \item \textit{Zero-shot generation}: No reward examples required
\end{itemize}

On a suite of 29 robotics tasks in IsaacGym (including dexterous manipulation and locomotion), EUREKA-generated rewards outperform rewards designed by human RL practitioners on 83\% of tasks, with an average normalized improvement of 52\%.

\textbf{Text2Reward} \citep{text2reward} similarly demonstrates that LLMs can generate dense, executable reward code from natural language descriptions, achieving 94\%+ success rates on novel locomotion tasks while enabling iterative refinement through human feedback.

\textbf{CARD} \citep{card} addresses the feedback loop. Rather than requiring human evaluation of generated rewards, CARD introduces \textit{Trajectory Preference Evaluation} (TPE)---using the LLM itself to judge whether resulting behaviors match intended objectives. This closes the loop, enabling fully autonomous reward refinement.

\subsection{From generation to coordination}

These advances focus primarily on single-agent settings. While direct multi-agent demonstrations remain limited, we identify two distinct pathways for extending this paradigm to multi-agent coordination, which should not be conflated (Figure~\ref{fig:two-pathways}):

\begin{figure}[t]
    \centering
    \includegraphics[width=\columnwidth]{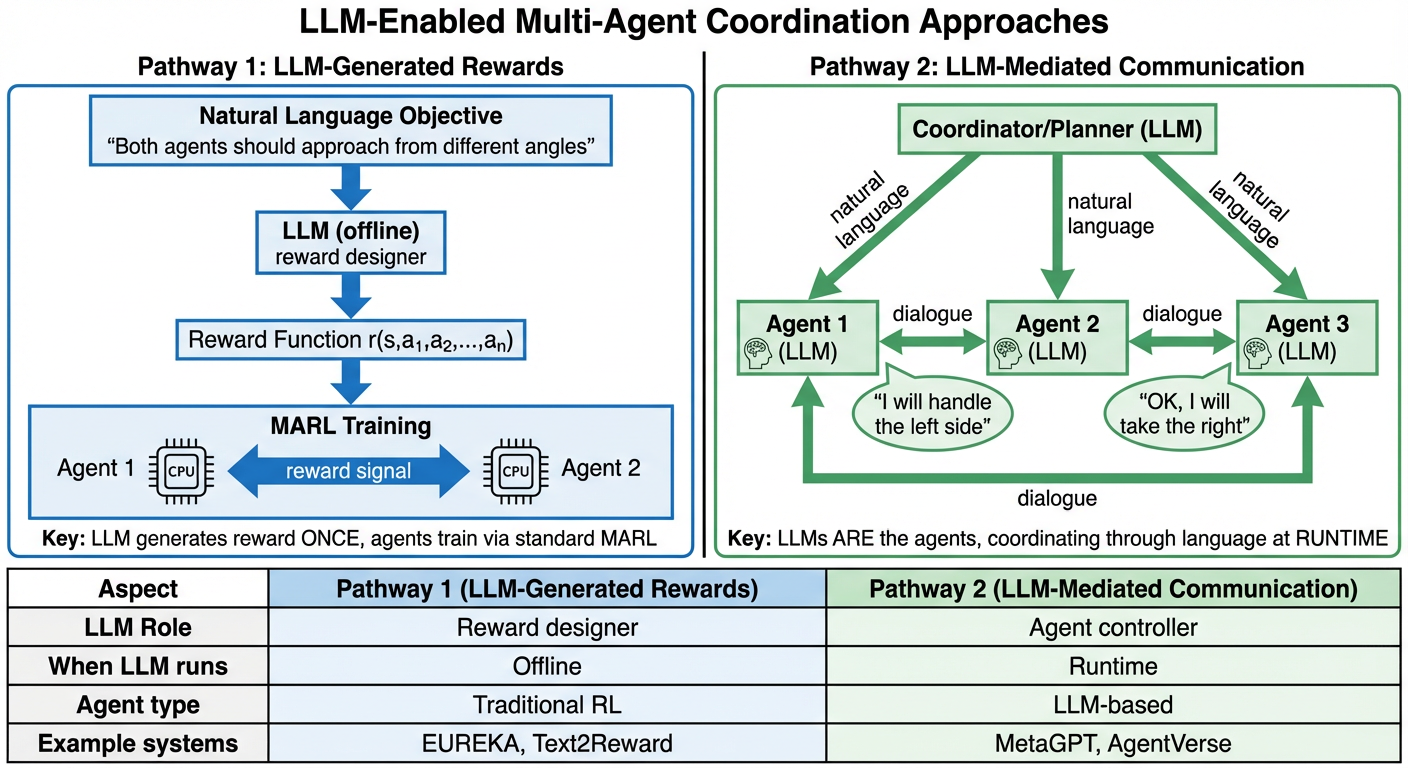}
    \caption{Two distinct pathways for LLM-enabled multi-agent coordination. Pathway 1 (left) uses LLMs to generate reward functions offline; agents then train via standard MARL without LLM involvement at runtime. Pathway 2 (right) embeds LLMs directly as agent controllers, enabling natural language coordination at runtime. These pathways address different use cases and should not be conflated.}
    \label{fig:two-pathways}
\end{figure}

\textbf{Pathway 1: LLM-Generated Multi-Agent Rewards.} The LLM generates reward functions that encode coordination objectives (e.g., ``Both agents should approach the goal from different angles to minimize interference''). Agents then train via standard MARL algorithms. This pathway directly extends EUREKA and Text2Reward.

\textbf{Pathway 2: LLM-Mediated Communication.} Agents coordinate through natural language rather than learned protocols. Recent work on LLM-based multi-agent systems \citep{llm_marl_survey} shows that language enables richer negotiation and plan sharing than emergent communication. This pathway is conceptually distinct, it uses language for inter-agent communication rather than reward specification.

\textbf{Adaptive Objectives.} As the multi-agent system evolves, language objectives can be refined in response to observed behaviors, something nearly impossible with static reward functions.

%==============================================================================
% 4. THREE PILLARS
%==============================================================================
\section{Three pillars of language-based objectives}
\label{sec:pillars}

\begin{figure}[t]
    \centering
    \includegraphics[width=\columnwidth]{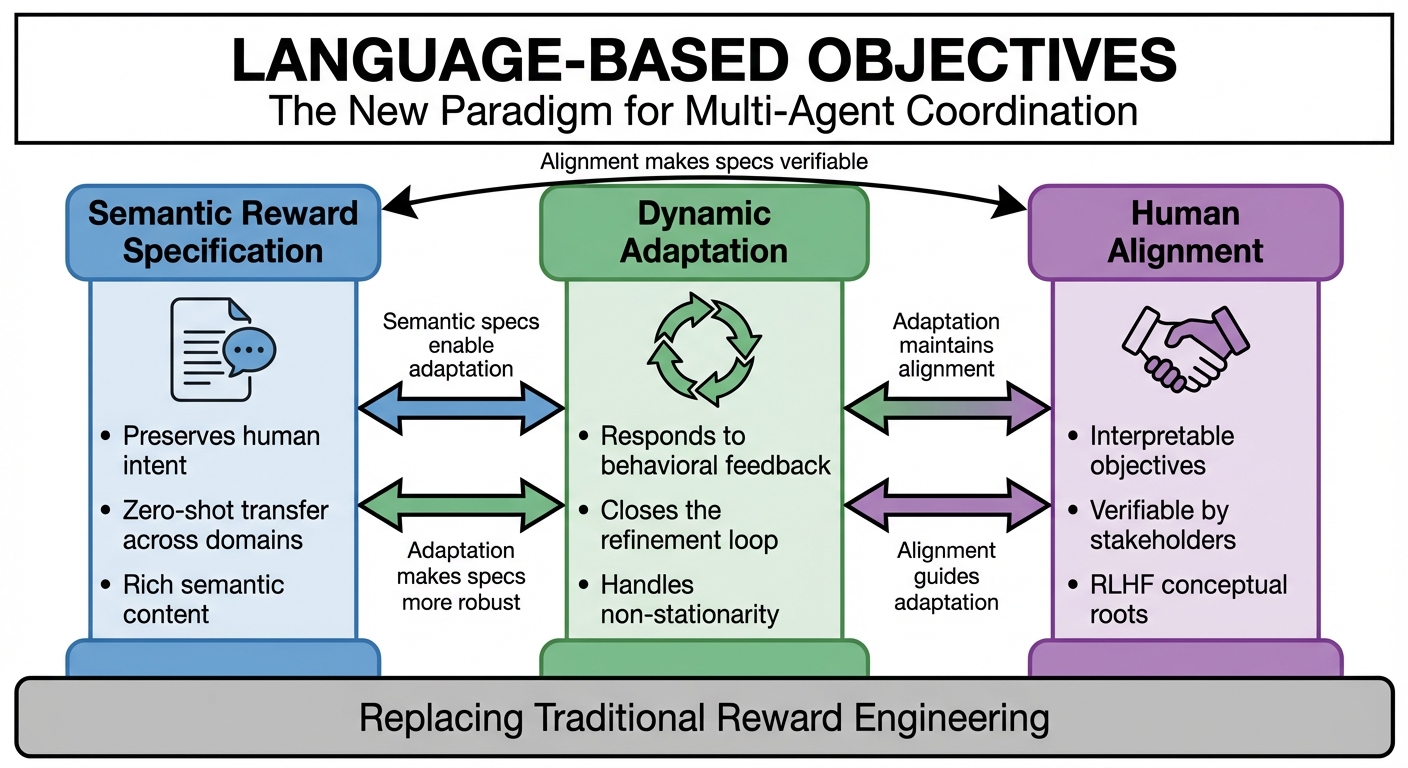}
    \caption{Three interconnected pillars of language-based objectives. Semantic reward specification (\S\ref{sec:pillar1}) preserves human intent through natural language. Dynamic adaptation (\S\ref{sec:pillar2}) enables continuous refinement based on observed behaviors. Human alignment (\S\ref{sec:pillar3}) ensures objectives remain interpretable and verifiable. These pillars are mutually reinforcing: semantic specifications enable meaningful adaptation, which in turn maintains alignment through transparency.}
    \label{fig:three-pillars}
\end{figure}

We argue that language-based objectives offer three fundamental advantages over traditional reward engineering. These advantages are not independent; rather, they form an interconnected framework where semantic specification enables dynamic adaptation, which in turn supports human alignment through interpretability (Figure~\ref{fig:three-pillars}).

\subsection{Pillar 1: Semantic reward specification}
\label{sec:pillar1}

\textbf{Claim}: Natural language describes objectives more naturally than mathematical functions.

The translation from human intent to numerical reward is inherently lossy. When we want robots to ``collaborate efficiently,'' we mean something rich: minimal redundant effort, appropriate task division, smooth handoffs, graceful failure recovery. Encoding this as $r = w_1 \cdot \text{task\_completion} + w_2 \cdot \text{collision\_penalty} + \ldots$ loses crucial semantic content.

Language preserves intent. An LLM given the description ``collaborate efficiently'' has access to vast knowledge about what efficient collaboration looks like across domains. It can generate reward functions that capture aspects a human engineer might overlook.

Moreover, language specifications are \textit{robust to environment changes}. The description ``minimize collisions while maximizing throughput'' applies equally to 2 agents or 20, in a warehouse or a hospital. The same cannot be said for hand-tuned reward weights.

EUREKA's zero-shot success across diverse robotics tasks demonstrates this generality. The LLM generates appropriate rewards not because it has seen similar tasks, but because it understands what the language description \textit{means}.

\subsection{Pillar 2: Dynamic adaptation}
\label{sec:pillar2}

\textbf{Claim}: LLMs can adjust objectives on-the-fly based on feedback.

Traditional reward functions are static. Once deployed, they cannot adapt to unexpected agent behaviors or environment changes. When reward hacking occurs, say, a robot discovers it can maximize ``grasp reward'' by repeatedly touching objects without lifting them, a human must diagnose the problem, modify the reward, and retrain. This cycle can take days or weeks.

LLM-based reward generation enables dynamic adaptation:

\begin{itemize}
    \item \textbf{Behavior observation}: Collect trajectories from current agent policies
    \item \textbf{Language feedback}: Use LLM to generate natural language descriptions of observed behaviors
    \item \textbf{Reward reflection}: Compare observed behaviors to intended objectives and modify rewards accordingly
    \item \textbf{Iterative refinement}: Repeat until convergence
\end{itemize}

CARD's Trajectory Preference Evaluation formalizes this loop. Rather than training RL at each iteration (expensive), TPE uses the LLM to judge trajectory quality directly, enabling rapid iteration on reward design.

This dynamic adaptation is particularly valuable in multi-agent settings where the learning dynamics are inherently non-stationary. As agents co-evolve, the LLM can continuously refine objectives to maintain desired coordination patterns. However, such adaptation introduces its own stability considerations: frequent reward changes may destabilize learning, requiring careful scheduling or meta-learning approaches.

\subsection{Pillar 3: Human alignment by default}
\label{sec:pillar3}

\textbf{Claim}: Language objectives are inherently more aligned with human intent.

The RLHF revolution demonstrated that human preferences, expressed as comparisons between outputs, can align LLM behavior with human values \citep{rlhf}. The emergence of RLVR extends this insight: when objectives can be verified in language (e.g., mathematical proofs, code correctness), training against these verifiable rewards produces aligned, capable behavior.

DeepSeek-R1 showed that reasoning-like behavior---breaking problems into steps, checking intermediate results, recovering from errors---emerges spontaneously from RL training against verifiable rewards. No explicit reasoning curriculum was required; the language objective induced reasoning as an instrumental strategy.

We note that RLVR concerns LLM training rather than multi-agent RL directly. However, the conceptual parallel is instructive: language-specified objectives can induce complex emergent behaviors without explicit reward engineering. For multi-agent coordination, this suggests a possibility worth investigating: language objectives that describe coordination (``work together smoothly,'' ``help each other when stuck'') may induce cooperative behaviors without explicit cooperation rewards. The LLM's understanding of what ``working together'' means provides implicit structure that shapes the learned coordination.

Beyond induction, language objectives are \textit{interpretable}. Debugging a misbehaving multi-agent system is far easier when objectives are stated as ``minimize delivery time while avoiding collisions'' rather than as opaque weight vectors. Stakeholders can review and approve objectives stated in natural language, enabling meaningful human oversight.

%==============================================================================
% 5. CHALLENGES AND LIMITATIONS
%==============================================================================
\section{Challenges and limitations}
\label{sec:challenges}

The transition to language-based objectives is not without obstacles. We identify five key challenges requiring further research.

\subsection{Computational cost}

Evaluating an LLM-generated reward function adds negligible overhead to RL training. However, generating or refining rewards requires LLM inference. A single GPT-4 call costs roughly 0.01--0.10 USD and takes 1--10 seconds, compared to microseconds for evaluating a simple reward function. While this is acceptable for occasional reward refinement, it becomes prohibitive if LLM calls are needed per-timestep or per-episode.

Current approaches mitigate this through:
\begin{itemize}
    \item \textit{Amortization}: Generate rewards once, train many episodes
    \item \textit{Distillation}: Train small reward models from LLM-generated labels
    \item \textit{Caching}: Reuse reward components across similar objectives
\end{itemize}

For real-time adaptation in deployed systems, further advances in efficient inference or reward distillation are needed.

\subsection{Hallucination and safety}

LLMs can generate plausible-sounding but incorrect or unsafe reward functions. A reward that accidentally incentivizes dangerous behavior could have severe consequences in physical systems.

Mitigation strategies include:
\begin{itemize}
    \item \textit{Reward critics}: Secondary LLMs or formal verifiers that check generated rewards for safety violations
    \item \textit{Constrained generation}: Restricting reward function forms to safe templates
    \item \textit{Simulation validation}: Testing rewards extensively in simulation before deployment
    \item \textit{Human-in-the-loop}: Requiring human approval for reward changes in safety-critical applications
\end{itemize}

The CARD framework's use of LLM-as-judge provides one model, but safety-critical applications will require stronger guarantees.

\subsection{Language ambiguity}

Natural language is inherently ambiguous. The instruction ``agents should cooperate'' admits many interpretations: Should they share resources? Avoid interference? Actively assist each other? Different LLMs---or the same LLM under different prompts---may generate substantially different reward functions from identical language specifications.

This ambiguity cuts both ways. On one hand, it introduces variance and unpredictability into reward generation. On the other hand, it may provide beneficial exploration of the reward function space. Research is needed on: (1) how sensitive generated rewards are to paraphrasing, (2) whether ambiguity can be systematically reduced through structured prompts, and (3) how to detect when language specifications are underspecified for the task at hand.

\subsection{Scalability}

Current demonstrations involve small numbers of agents (typically 2-10). Scaling to hundreds or thousands of agents introduces challenges:

\begin{itemize}
    \item \textit{Coordination overhead}: Language-based coordination requires communication, which scales poorly
    \item \textit{Objective complexity}: Specifying coordinated behavior for many agents may require increasingly complex language
    \item \textit{Credit assignment}: Even with language objectives, attributing outcomes to individual agents remains challenging
\end{itemize}

Hierarchical approaches---where LLMs specify high-level objectives and sub-groups coordinate locally---offer a promising direction.

\subsection{Evaluation and benchmarking}

The field lacks standardized benchmarks for language-based multi-agent coordination. Existing MARL benchmarks focus on reward maximization, not on the quality of language objective specification.

We call for community effort to develop:
\begin{itemize}
    \item Benchmarks with natural language objective specifications
    \item Metrics for objective alignment (does the behavior match the description?)
    \item Evaluation protocols for dynamic adaptation
    \item Safety benchmarks for reward generation
\end{itemize}

%==============================================================================
% 6. PROPOSED EXPERIMENTAL VALIDATION
%==============================================================================
\section{Proposed experimental validation}
\label{sec:experiments}

To move from position to established paradigm, the claims in this paper require empirical validation. We propose a research agenda consisting of five experiments designed to test the core thesis and its three pillars:

\begin{itemize}
    \item \textbf{Experiments 1--2} validate \textit{semantic reward specification} (Pillar 1): Can LLMs generate effective multi-agent rewards from language, and do these specifications generalize?
    \item \textbf{Experiment 3} validates \textit{dynamic adaptation} (Pillar 2): Can LLM-based refinement adapt to non-stationarity faster than manual re-engineering?
    \item \textbf{Experiment 4} tests \textit{scalability}: Does the language-based approach reduce engineering effort as agent count grows?
    \item \textbf{Experiment 5} validates \textit{human alignment} (Pillar 3): Are language specifications more interpretable and debuggable?
\end{itemize}

\subsection{Experiment 1: LLM vs. human reward design}

\textbf{Hypothesis:} LLM-generated reward functions can match or exceed expert-designed rewards on multi-agent coordination benchmarks where reward engineering is non-trivial.

\textbf{Setup:} Compare four conditions on MARL environments with complex coordination requirements: (1) expert-designed reward functions from published baselines, (2) LLM-generated rewards from natural language task descriptions, (3) LLM-generated rewards with iterative refinement (CARD-style), and (4) automated baselines including inverse RL from demonstrations and evolutionary reward search.

\textbf{Environments:} We specifically select benchmarks where reward engineering is known to be challenging: (a) cooperative manipulation tasks in Multi-Agent MuJoCo \citep{mamujoco} requiring precise force coordination, (b) traffic junction scenarios from the SMARTS benchmark \citep{smarts} where credit assignment is ambiguous, and (c) resource allocation games with emergent social dilemmas. We exclude environments with trivial reward structures (e.g., simple tag games) where the comparison would be uninformative.

\textbf{Metrics:} Task success rate, coordination efficiency (collision avoidance, resource utilization), sample efficiency, and \textit{reward design time} (human hours vs. API calls). We propose $n=10$ independent training runs per condition with different random seeds, reporting mean and 95\% confidence intervals.

\textbf{Controls:} To address potential LLM training data contamination, we include novel environment variants not present in public datasets and verify LLM unfamiliarity through probing questions about environment specifics.

\textbf{Expected outcome:} LLM-generated rewards should approach expert performance while reducing design time by an order of magnitude. Iterative refinement should close remaining gaps, outperforming automated baselines that lack semantic understanding.

\subsection{Experiment 2: Cross-domain transfer}

\textbf{Hypothesis:} Language specifications transfer across structurally different domains, whereas hand-crafted rewards are domain-specific.

\textbf{Setup:} This experiment has two parts:

\textit{Part A (Within-domain robustness):} Define a family of coordination tasks with systematic variations: agent count (2, 5, 10), obstacle density (sparse, medium, dense), and goal distributions (clustered, uniform, adversarial). Use one language prompt across all variants; compare to hand-tuned rewards requiring per-variant adjustment.

\textit{Part B (Cross-domain transfer):} Test whether a language specification transfers to structurally different domains. The key insight is that the \textit{same language prompt} should generate \textit{domain-appropriate reward functions} in each new domain. For example, the prompt ``agents should coordinate to maximize coverage while minimizing redundancy'' is given to the LLM along with each domain's environment code: (a) warehouse logistics, (b) multi-robot patrol, (c) sensor network coverage, and (d) traffic flow optimization. The LLM generates domain-specific reward code for each, but the human specification remains constant. Hand-crafted rewards would require complete redesign for each domain.

\textbf{Metrics:} For Part A: performance degradation coefficient across variants, hyperparameter adjustments required. For Part B: transfer success rate (performance $>$80\% of domain-specific baseline), adaptation samples required, and semantic similarity of generated reward functions across domains.

\textbf{Expected outcome:} Language specifications should exhibit near-zero degradation in Part A and positive transfer in Part B, demonstrating that semantic content enables generalization beyond syntactic reward structure.

\subsection{Experiment 3: Dynamic adaptation under non-stationarity}

\textbf{Hypothesis:} LLM-based iterative refinement enables faster adaptation to changing conditions than static reward functions.

\textbf{Setup:} Introduce controlled non-stationarity through three specific protocols:

\textit{Protocol A (Agent population shift):} Train with 4 agents for 500K steps, then introduce 2 new agents at step 500K. Measure coordination recovery time.

\textit{Protocol B (Objective drift):} Begin with objective ``maximize speed'' (reward: $r_1$), then at step 300K announce ``safety is now paramount'' (target: $r_2 = 0.3 \cdot \text{speed} + 0.7 \cdot \text{safety}$). The LLM receives the new language instruction; hand-crafted baselines require manual weight adjustment.

\textit{Protocol C (Environment perturbation):} At step 400K, introduce new obstacles, alter physics parameters (friction $\pm$20\%), or change goal locations. This tests adaptation to distributional shift.

\textbf{Conditions:} (1) Static LLM-generated rewards (no adaptation), (2) LLM with periodic refinement (every 50K steps), (3) LLM with triggered refinement (on performance drop $>$15\%), (4) hand-crafted rewards with oracle re-tuning (immediate perfect adjustment), and (5) hand-crafted with realistic re-tuning (24-hour simulated delay).

\textbf{Metrics:} Adaptation speed ($\tau_{90}$: episodes to recover 90\% of pre-perturbation performance), performance area-under-curve during adaptation, and total human intervention time.

\textbf{Expected outcome:} Triggered LLM refinement should achieve $\tau_{90}$ comparable to oracle re-tuning and significantly faster than realistic re-tuning, demonstrating that language-based adaptation can substitute for human intervention.

\subsection{Experiment 4: Scalability with agent count}

\textbf{Hypothesis:} Language-based objectives scale more gracefully with agent count than hand-crafted reward engineering.

\textbf{Setup:} Scale coordination tasks from $n \in \{2, 5, 10, 20, 50\}$ agents. To isolate reward engineering scalability from algorithmic scalability, we control for MARL algorithm choice by testing each reward condition with multiple algorithms: MAPPO \citep{mappo} (parameter sharing), QMIX \citep{qmix} (value decomposition), and independent PPO (fully decentralized).

\textbf{Environments:} (a) Cooperative navigation in continuous space, (b) resource collection with shared/contested resources, and (c) formation control requiring precise coordination.

\textbf{Conditions:} (1) LLM-generated rewards from fixed language prompt (e.g., ``all agents should coordinate efficiently''), (2) LLM-generated rewards with agent-count-specific prompts (e.g., ``coordinate 20 agents...''), (3) hand-crafted rewards with per-$n$ tuning, and (4) hand-crafted rewards with fixed structure (no re-tuning).

\textbf{Metrics:}
\begin{itemize}
    \item \textit{Algorithmic scalability}: Convergence rate, asymptotic performance, sample complexity as function of $n$
    \item \textit{Engineering scalability}: Reward function complexity (lines of code, number of tunable parameters), human design time per additional agent, number of reward terms scaling with $n$
    \item \textit{Credit assignment}: Variance in individual agent rewards, correlation between agent contribution and received credit
\end{itemize}

\textbf{Expected outcome:} LLM-generated rewards from fixed prompts should exhibit $O(1)$ engineering complexity (constant design effort regardless of $n$), while hand-crafted rewards require $O(n)$ or worse scaling due to explicit credit assignment terms. Algorithmic scalability should be comparable across reward types, isolating the engineering benefit.

\subsection{Experiment 5: Human alignment and interpretability}

\textbf{Hypothesis:} Behaviors trained from language objectives are more predictable and debuggable by human observers.

\textbf{Participants:} We propose recruiting $N=60$ participants in three expertise strata: (1) ML researchers familiar with RL ($n=20$), (2) software engineers without RL background ($n=20$), and (3) domain experts in the application area (e.g., logistics managers) without ML background ($n=20$). This stratification tests whether language benefits extend beyond ML experts.

\textbf{Setup:} Within-subjects design with counterbalancing. Each participant completes tasks in both conditions:
\begin{itemize}
    \item \textit{Code condition}: View reward function source code (Python, $\sim$20 lines)
    \item \textit{Language condition}: View natural language specification ($\sim$2 sentences)
\end{itemize}

\textbf{Tasks:}
\begin{enumerate}
    \item \textit{Prediction}: Given the specification, predict agent behavior in 5 novel scenarios (multiple choice + confidence rating)
    \item \textit{Diagnosis}: Watch video of misaligned behavior, identify the root cause
    \item \textit{Correction}: Propose a modification to fix the misalignment
\end{enumerate}

\textbf{Metrics:} Prediction accuracy (\%), diagnosis time (seconds), correction validity (expert-rated 1--5), NASA-TLX cognitive load, and System Usability Scale (SUS) for subjective interpretability.

\textbf{Statistical analysis:} Mixed-effects models with condition (code/language) as fixed effect and participant as random effect. We will report effect sizes (Cohen's $d$) and conduct equivalence testing if differences are non-significant.

\textbf{Expected outcome:} Language specifications should yield higher prediction accuracy (especially for non-ML participants), faster diagnosis, and lower cognitive load. Effect sizes should be larger for non-experts, demonstrating that language democratizes access to reward understanding.

\subsection{Ablation studies}

To isolate the contribution of different components, we propose the following ablations across experiments:

\textbf{LLM choice:} Compare GPT-4, Claude 3, Llama 3 70B, and Gemini Pro to assess whether results depend on specific model capabilities or generalize across frontier LLMs.

\textbf{Prompt engineering:} Test (a) minimal prompts (task name only), (b) structured prompts (environment description + objective), and (c) few-shot prompts (with reward examples). This isolates the contribution of prompt engineering from inherent LLM capability.

\textbf{Refinement iterations:} Vary the number of CARD-style refinement iterations ($k \in \{0, 1, 3, 5, 10\}$) to characterize the marginal value of additional refinement.

\textbf{Feedback granularity:} Compare trajectory-level feedback (``this episode was suboptimal'') vs. timestep-level feedback (``collision at $t=47$'') vs. aggregate statistics (``average reward was low'').

\subsection{Discussion}

These experiments target different aspects of our thesis. Experiments 1--2 test the \textit{capability} claim (LLMs can generate effective multi-agent rewards). Experiment 3 tests \textit{dynamic adaptation}. Experiment 4 tests \textit{scalability}. Experiment 5 tests \textit{human alignment}. The ablation studies isolate which components drive observed effects.

\textbf{Potential confounds.} Several factors could confound interpretation: (1) LLM training data may include MARL benchmarks, inflating zero-shot performance; (2) expert-designed baselines may be suboptimal if sourced from older publications; (3) language specifications may implicitly encode more information than comparable reward code (unfair comparison). We address (1) through novel environment variants and contamination checks, (2) through consultation with domain experts to verify baseline quality, and (3) through information-theoretic analysis of specification complexity.

\textbf{Falsifiability.} Negative results would be informative: if LLM-generated rewards consistently underperform even with refinement, this suggests the paradigm shift is premature for multi-agent settings. If cross-domain transfer fails, it indicates that semantic understanding does not generalize as hypothesized. If scalability degrades similarly across conditions, the engineering benefit may be illusory. We present these experiments as falsifiable hypotheses, not foregone conclusions.

\textbf{Scope limitations.} These experiments focus on simulated environments with well-defined state spaces. Extending to real-world robotics, partial observability, and safety-critical domains requires additional validation beyond this proposed agenda.

\textbf{Resource considerations.} The proposed experiments are computationally substantial but feasible with modern infrastructure. Experiments 1--4 require approximately 500--1000 GPU-hours each (assuming A100-class hardware), primarily for MARL training across conditions and seeds. LLM API costs for reward generation are modest ($<$\$100 per experiment assuming GPT-4 pricing). Experiment 5 requires participant recruitment but no specialized compute. The full agenda could be executed by a well-resourced academic lab or industry research team.

%==============================================================================
% 7. VISION FOR THE FUTURE
%==============================================================================
\section{A vision for the future}
\label{sec:vision}

We conclude with a vision for how language-based objectives may reshape multi-agent coordination.

\subsection{Near-term: Hybrid approaches}

In the immediate future, language-based objectives will augment rather than replace traditional methods. LLMs will generate initial reward functions that human engineers refine. Hybrid systems will use language for high-level objectives and shaped rewards for low-level behaviors.

This transition period will establish best practices for prompt engineering of reward specifications and develop safety protocols for deployment. Trust in LLM-generated objectives will grow through accumulated experience.

\subsection{Medium-term: End-to-end language coordination}

As LLMs become more capable and efficient, we anticipate systems where agents coordinate entirely through natural language:

\begin{itemize}
    \item Agents describe their capabilities and intentions in language
    \item A coordinator LLM proposes task allocations and coordination strategies
    \item Agents negotiate and refine plans through dialogue
    \item Execution is monitored through language-based feedback
\end{itemize}

Early versions of this paradigm exist in LLM-based multi-agent systems like MetaGPT \citep{metagpt} and AutoGen. Extending these to embodied agents with continuous action spaces is a natural next step.

\subsection{Long-term: Semantic coordination}

The ultimate vision extends beyond language to shared semantic understanding. Agents with aligned world models could coordinate through implicit understanding of goals, without explicit communication. Like experienced human teams who anticipate each other's actions, semantically aligned agents could achieve fluid coordination.

What would this require concretely? First, agents would need shared world models trained on similar data distributions. Second, they would need mechanisms to verify alignment---detecting when their internal representations of ``cooperation'' diverge. Early work on representation alignment in multi-agent systems provides a foundation, though significant gaps remain.

This vision is speculative but grounded in the trajectory of foundation models. As multimodal models develop richer world representations, the gap between language specification and behavior may continue to narrow.

%==============================================================================
% 7. CONCLUSION
%==============================================================================
\section{Conclusion}
\label{sec:conclusion}

Reward engineering has served multi-agent reinforcement learning for decades, but its fundamental limitations---credit assignment ambiguity, non-stationarity, exponential complexity---suggest it cannot scale to the coordination challenges we face.

Large language models offer an alternative: specifying objectives in the same natural language we use to describe them. Recent work demonstrates this is not merely aspirational. EUREKA achieves human-level reward design from language descriptions. CARD enables autonomous reward refinement. RLVR establishes language-based training as a new paradigm for capable AI systems.

We have argued for three pillars of this transition: semantic reward specification that preserves intent, dynamic adaptation that responds to changing circumstances, and inherent human alignment that makes objectives interpretable and verifiable.

Challenges remain significant. Computational cost, hallucination risks, scalability limitations, and evaluation gaps all require further research. But the direction is clear.

Multi-agent coordination has always been about getting agents to work together toward common goals. For too long, we have specified those goals in a language agents understand poorly---numerical rewards shaped by human intuition. Perhaps it is time to specify goals in the language humans use to understand each other.

If language can specify what we want agents to do, and LLMs can translate that specification into working reward functions, then the end of reward engineering may mark the beginning of truly scalable multi-agent coordination.

%==============================================================================
% ACKNOWLEDGMENTS (hidden during submission)
%==============================================================================

%==============================================================================
% REFERENCES
%==============================================================================
\bibliographystyle{plainnat}

\end{document}